\documentclass{article} 
\usepackage{iclr2027_conference,times}

\usepackage{amsmath,amsfonts,bm}

\def\eqref#1{equation~\ref{#1}}

\def\1{\bm{1}}

\DeclareMathAlphabet{\mathsfit}{\encodingdefault}{\sfdefault}{m}{sl}
\SetMathAlphabet{\mathsfit}{bold}{\encodingdefault}{\sfdefault}{bx}{n}

\usepackage{hyperref}
\usepackage{url}

\usepackage{wrapfig}
\usepackage{booktabs}
\usepackage{xspace}
\usepackage{multirow}
\usepackage{graphicx}
\usepackage{subcaption}
\usepackage{comment}
\newcommand{\modelname}{\texttt{\textbf{VERITAS}}}

\title{Verbalising Model's Computation for \\ Chain-of-Thought Faithfulness Detection}

\author{
Yujing Wang$^{1}$ \quad Yulan He$^{1}$ \quad Wenya Wang$^{2}$ \quad Hanqi Yan$^{1}$ \\
$^{1}$King's College London \quad $^{2}$Nanyang Technological University \\
\texttt{\{yujing.3.wang,yulan.he,hanqi.yan\}@kcl.ac.uk}, \ \texttt{wangwy@ntu.edu.sg}
}

\iclrfinalcopy
\begin{document}

\maketitle
\lhead{Preprint}

\begin{abstract}

Current Chain-of-Thought (CoT) verification methods detect reasoning faithfulness from model outputs or static activations, but are limited in explaining how unfaithfulness develops during generation. We introduce \modelname{} to verbalise the model's computation: how different reasoning anchors emerge, transfer across layers, and surface in the output. \modelname{} adopts the layer-wise Jacobian projection to produce a verbalised account of the model's internal reasoning evolution at each generated CoT token. By training a classifier on the extracted features along the traces, we show that these structured features provide insightful perspectives on how and where reasoning diverges toward unfaithfulness. Experiments on four datasets from FaithCoT-Bench and BonaFide show that \modelname{} achieves a significant performance gain over LLM-as-Judge, with a 28.4\% gain in balanced accuracy, and an average improvement of 16.0\% over activation probing. Further analysis shows \modelname{} can be generalised to different datasets and model architectures. Our work shows that verbalising model computation provides a richer and more interpretable lens on CoT faithfulness, moving beyond detection toward a mechanistic understanding of how reasoning fails.

\end{abstract}

\section{Introduction}
Chain-of-thought (CoT) improves the LLMs' reasoning efficiency and explainability \citep{wei2022chain,kojima2022large} by externalising the model's thinking. However, recent evidence shows that \textit{''CoTs are not faithful''} and can be affected by misleading cues, concealing the reward-hacking problem~\citep{turpin2023language,paul2024making,wang2026thinking}.

Existing approaches detect unfaithfulness either from model outputs or from static activation snapshots. LLM-as-a-Judge evaluates whether the visible CoT plausibly supports its answer, but cannot access the model's internal computation \citep{shen2026faithcot,yazdani2026comprehensive}. Activation-based methods probe whether answer-relevant information is causally mediated through the CoT \citep{yeo2025towards,kudo2026llms,jia2026faithfulness}, and linear probes show that fixed-layer hidden activations can expose deceptive behaviour \citep{goldowsky2025detecting}. Yet none of these approaches explain how unfaithfulness develops during generation, as they do not model how reasoning-relevant information emerges, transfers across layers, and is reflected in the generated CoT.

To address this gap, we focus on verbalising the model's computation, tracing how reasoning-relevant information arises, propagates through layers, and shapes each generated CoT token. Unlike prior approaches that capture a static snapshot of model behaviour, this yields a continuous, mechanistic account of how reasoning evolves throughout generation. We hypothesize that faithful and unfaithful CoTs induce distinct internal trajectories in this evolution, and that by tracing these trajectories we can not only detect unfaithfulness but explain where and how reasoning diverges.

To verify this hypothesis, we propose \modelname{} (\textbf{VER}balise the \textbf{I}nternal computation \textbf{T}race for F\textbf{A}ithfulne\textbf{S}s) to detect the reasoning divergence via comparison against transitional faithful anchors. Specifically, we leverage the J-lens~\citep{gurnee2026verbalizable} to map developmental hidden states across the model's layers into a common, interpretable representation space, where the anchor tokens are adopted to verbalise significant evolution stages. To establish faithfulness anchors, we take the differences between a paired faithful and unfaithful CoTs and extract the two-dimensional evolution over the reasoning stage tokens. For a given evaluated CoT, we model its generation process by extracting its activated information across different model layers and different CoT phases. By aligning the structured features along the generation traces with faithfulness anchors, we can therefore verify whether and where the reasoning diverges. To consolidate these structured features, we train a classifier for detection, and the evaluation results show that it is not only efficient for classification but also provides a mechanistic explanation of unfaithful reasoning.


Our main contributions can be summarized as follows:
\begin{itemize}
    \item We introduce \modelname{}, a framework that verbalises model computation to produce a continuous, mechanistic account of how reasoning evolves during CoT generation.
    \item We design three structured trace features: \emph{Map}, \emph{Profile}, and \emph{Dynamics} to jointly characterise where anchor evidence appears, which anchor tokens are expressed, and how anchor states evolve across model layers and CoT phases.
    \item We show that a classifier trained on these features achieves a 28.4\% gain over LLM-as-Judge and 16.0\%\ over activation probing, while providing mechanistic explanations of reasoning failures with consistent generalisation across datasets and model architectures.
\end{itemize}
\section{Verbalise the CoT Internal Computation}
\label{sec:prelim}
Recent research on CoT monitoring either uses LLMs-as-Judge~\citep{mittal2026c2} to examine verbal cues or projects generated CoTs into embedding spaces to explore their semantics. Both lines of work analyze the generated CoTs post hoc, and limited attention has been paid to how a CoT develops through the model's internal computation. In this section, we introduce how to read out CoT development along the model's internal computation. This lays the technical foundation for the subsequent faithfulness detection with transitional features along two axes.

\subsection{Understanding model computation via Jacobian lens}
To understand how information flows throughout model layers, we adopt the Jacobian lens~\citep{gurnee2026verbalizable} to explain layer-wise information by associating it with the vocabulary tokens. 

With $\mathbf{h}_{\ell,t}\in\mathbb{R}^{d}$ for the residual stream at layer $\ell$ and token position $t$, the Jacobian lens summarizes how a state at an intermediate layer influences what the model later says, by linearizing the map from that state to the final-layer residual. For a source layer $\ell$, its lens matrix is: 
\begin{equation}
\mathbf{J}_\ell \;=\; \mathbb{E}_{t}\!\left[\, \sum_{t' \geq t} \frac{\partial \mathbf{h}_{L,t'}}{\partial \mathbf{h}_{\ell,t}} \,\right],
\label{eq:jacobian}
\end{equation}
where $L$ is the final residual block. Following \citet{gurnee2026verbalizable}, we estimate the lens by averaging the Jacobian over a generic corpus. Applying the lens to an activation first transports it to the final-layer residual space and then applies the model's normal unembedding:
$\operatorname{lens}_{\ell}(\mathbf{h}_{\ell,t})=\operatorname{softmax}\!\left(\mathbf{W}_U\,\operatorname{norm}\!\left(\mathbf{J}_{\ell}\mathbf{h}_{\ell,t}\right)\right),
\label{eq:jlens}$
where $\mathbf{W}_U\in\mathbb{R}^{|V|\times d}$ is the unembedding matrix and $\operatorname{norm}$ is the model's final normalization. This produces a vocabulary distribution describing what the intermediate state makes the model express.

The rows of the matrix $\mathbf{D}_{\ell} = \mathbf{W}_U\mathbf{J}_{\ell}$ are the J-lens vectors at layer $\ell$. We denote the row for token $w$ by $\tilde{\mathbf{j}}_{\ell,w}=\left(\mathbf{W}_U\mathbf{J}_{\ell}\right)[w,:]$, which is a direction in the layer-$\ell$ space associated with the model's disposition to express its corresponding token. The collection of J-lens vectors forms \emph{J-space}.  


\subsection{Readouts along depth and reasoning progress}
\label{sec:readouts}

With access to model internal computation,  we now compare how CoT develops along the model depth axis and the external generation steps.

Given a question $q$, the visible CoT is a verbalization of this internal information flow and may omit or distort information that influenced the answer \citep{turpin2023language}. The model's computation is instantiated in its evolving internal states $\mathbf{h}_{\ell,t}$, which carry reasoning-relevant information across layers and generation steps. We therefore view faithfulness as a relation between two processes: the propagation of reasoning-relevant information through the model's internal states and its progressive expression in the visible CoT. Our central hypothesis is that faithful and unfaithful reasoning induce different semantic trajectories along both model depth and reasoning progress.

We sample residual states $\mathbf{h}_{\ell,t}\in\mathbb{R}^{d}$ at sentence and clause boundaries and group them into $P$ normalized CoT phases, 
$\bar{\mathbf{h}}_{\ell,p}=\frac{1}{|\mathcal{T}_p|}\sum_{t\in\mathcal{T}_p}\mathbf{h}_{\ell,t},$
where $\mathcal{T}_p$ contains the readouts in phase $p$. This yields a fixed layer-by-phase grid for every trace.

\paragraph{Following semantics across reasoning steps.}
At each layer, we trace the J-space alignment across normalized CoT phases. To compare the internal state with the visible CoT, let $\mathbf{f}_p$ denote its normalized token-frequency vector in phase $p$. Using the same layer-specific J-space dictionary $\mathbf{D}_\ell$, CoT can be transformed in J-space as $\mathbf{z}^{\mathrm{vis}}_{\ell,p}
=\mathbf{D}_\ell\operatorname{norm}_2(\mathbf{D}_\ell^\top\mathbf{f}_p))$. The alignment vectors $\mathbf{z}^{\mathrm{in}}_{\ell,p}$ and $\mathbf{z}^{\mathrm{vis}}_{\ell,p}$ share the same token coordinates and represent internally expressed and explicitly verbalized content, respectively.
\begin{figure*}[t]
\centering
\includegraphics[width=0.7\textwidth]{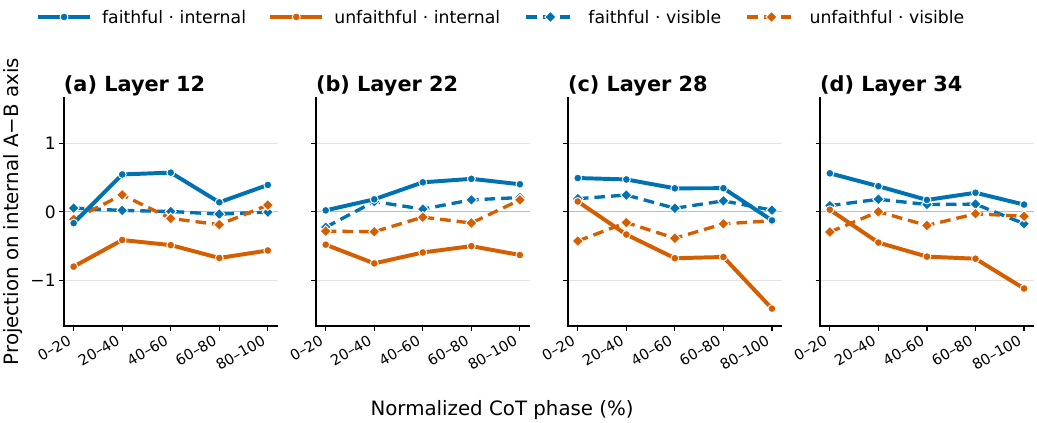}
\caption{Internal (solid) and visible-CoT lexical-proxy (dashed) trajectories on TruthfulQA, projected onto the internal faithful--unfaithful J-space direction. }
\label{fig:internal-visible-trajectories}
\end{figure*}

Figure~\ref{fig:internal-visible-trajectories} projects both trajectories onto the faithful-minus-unfaithful direction defined by J-space distributions. The faithful internal and visible trajectories also exhibit more similar phase-wise trends, particularly near the end of the CoT. This pattern is consistent with faithful verbalizations preserving the evolution of internal semantic content more accurately.

\paragraph{Following semantics across layers.}
We project each phase-level state $\bar{\mathbf{h}}_{\ell,p}$ into the J-space vectors defined in~\S\ref{sec:prelim}, obtaining $\mathbf{z}^{\mathrm{in}}_{\ell,p}=\mathbf{D}_{\ell}\bar{\mathbf{h}}_{\ell,p}/\lVert\bar{\mathbf{h}}_{\ell,p}\rVert$. Because each coordinate denotes the same token across layers, $\{\mathbf{z}^{\mathrm{in}}_{\ell,p}\}_{\ell\in\mathcal{L}}$ forms a depth-wise semantic trajectory in a shared lexical space.

\begin{figure*}[t]
\centering
\includegraphics[width=0.8\textwidth]{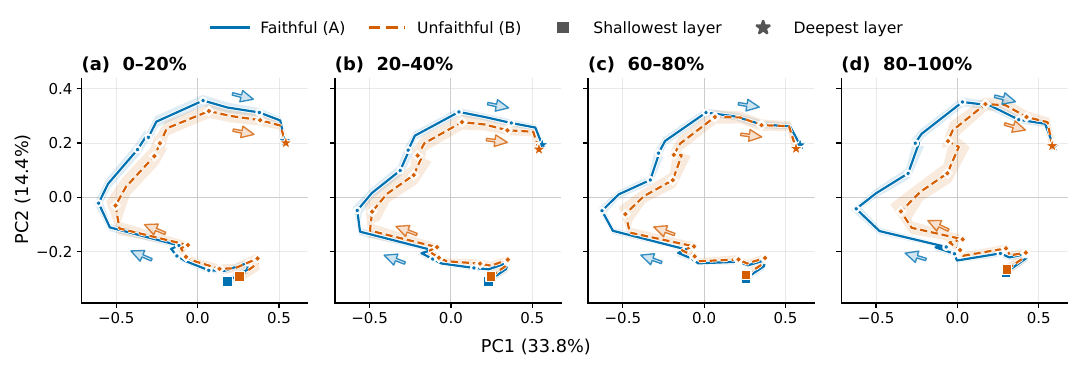}
\caption{J-space trajectories across model depth at four normalized CoT phases on AQuA. Panels (a)--(d) follow the progression of the visible CoT. Arrows indicate increasing layer depth, and shading shows approximate 95\% confidence intervals. 
}
\label{fig:jspace-layer-trajectories}
\end{figure*}

Figure~\ref{fig:jspace-layer-trajectories} shows how the J-space representation evolves across model depth at different points in the CoT. Panels progress from the first 20\% to the final 20\% of the normalized CoT, while each trajectory runs from shallow to deep layers. Faithful and unfaithful traces follow a similar global path through depth, but exhibit systematic, phase-dependent offsets. Their separation is clearest near the middle layers in the final CoT phase, indicating that faithfulness is reflected in how semantic content moves through the network rather than at one isolated layer.

\section{Tracing CoT Dynamics for Faithful Detection}
\label{sec:method}
We argue that \textit{``CoT is faithful when its visible rationale consistently reflects how reasoning-relevant information is computed in the model's internal decision-making process''.} Using computation readout techniques, we detail how to detect unfaithfulness from its transitional dynamics.


\subsection{Overview}

Given a query $q$ and a generated CoT $c$, let
$\mathcal{H}(q,c)=\{\mathbf{h}_{\ell,t}\}_{\ell,t}$ denote the corresponding internal states. Our goal is to predict a faithfulness label $y\in\{A,B\}$, where $A$ and $B$ denote faithful and unfaithful CoTs, respectively. We extract a feature vector $\mathbf{x}=\Phi(\mathcal{H}(q,c))$ from the internal states and learn a detector. Unlike previous probes~\citep{goldowsky2025detecting}, we build $\Phi$ with J-lens to verbalize internal states through aligned anchor tokens.

\begin{figure*}[t]
\centering
\includegraphics[width=0.8\textwidth]{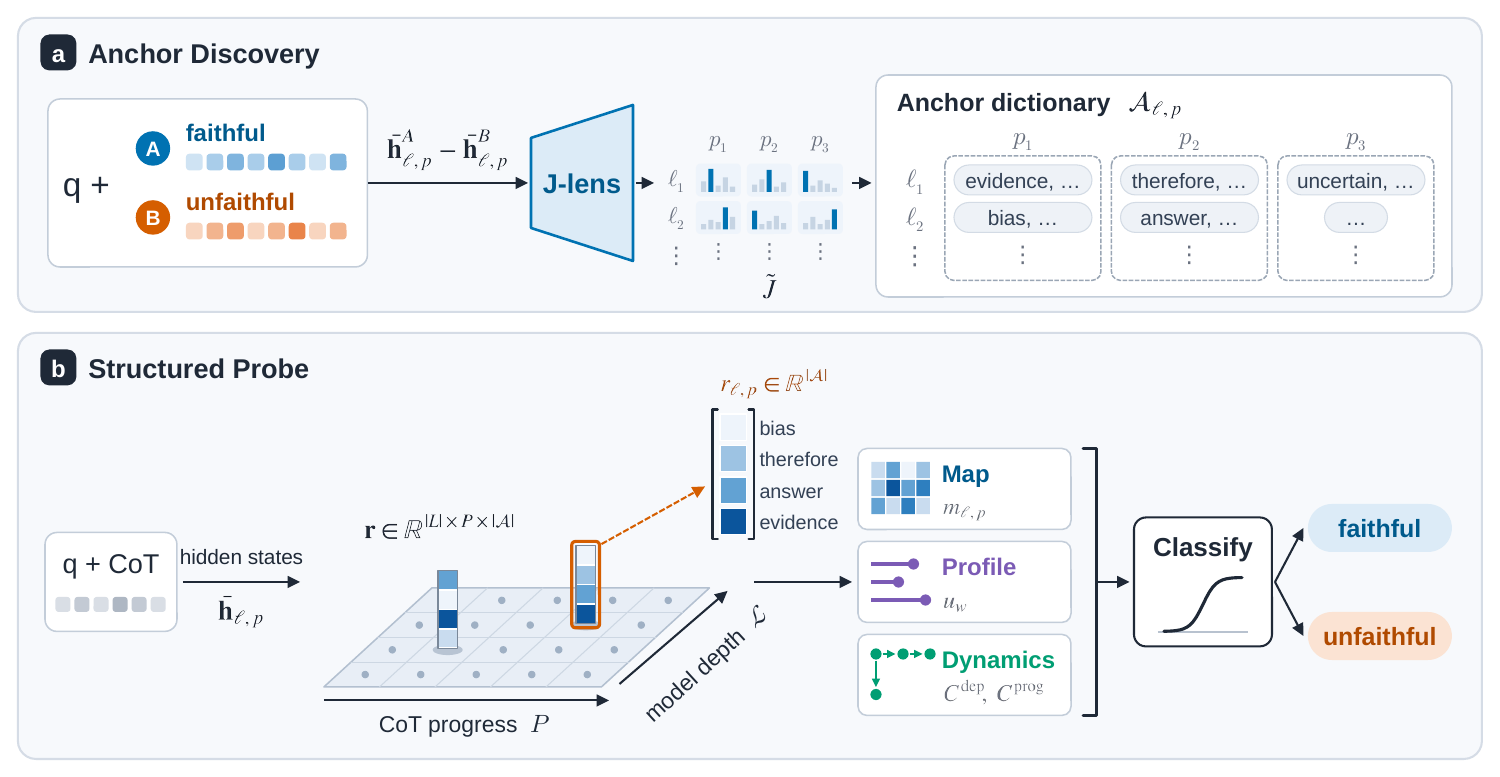}
\vspace{-1mm}
\caption{Overview. \textbf{Top}: A contrast CoT pair is mapped to a shared J-space for faithful anchor ($\tilde{J}$) discovery. The resulting anchor dictionary $\mathcal{A}_{\ell,p}$ explains how reasoning stages shift across different model layers $\ell$ as CoT progression ($p$) via representative tokens. \textbf{Bottom}: Given an evaluation CoT, the structured faithful detector is built to measure the CoT's transitional features (Map, Profile, Dynamics) against the discovered faithful anchor.}
\label{fig:overview}
\end{figure*}

Figure~\ref{fig:overview} illustrates our method. Raw hidden states mix faithfulness with nuisance factors such as style and task content, and they are not comparable across layers. Therefore, we firstly discover the faithful \textit{\textbf{anchor}}, i.e., J-lens vectors,$\tilde{J}$ that describe the reasoning-informative flows across different model layers as the CoT phase progresses. (\S\ref{sec:discovery}). 
Since J-lens maps all layers into a shared token space, these anchors provide a common, interpretable coordinate system, called \textbf{\textit{Anchor Dictionary}} $\mathcal{A}$. We adopt the J-lens vector to score each hidden state of a new CoT against its directions, yielding an alignment tensor over model depth and CoT progress (\S\ref{sec:views}). We propose three transitional features, i.e., \emph{Map}, \emph{Profile}, and \emph{Dynamics} to trace \emph{where} evidence appears, \emph{which} anchors are expressed, and \emph{how} the tensor evolves. A classifier $f_\theta(\mathbf{x})=P(y=A\mid\mathbf{x})$ detects faithfulness from the concatenated features $\mathbf{x}$.

\subsection{Constructing the Explainable Anchor}
\label{sec:discovery}

Anchor discovery proceeds in two steps. We first compute a faithful--unfaithful contrast direction at each layer--phase cell, and then translate it into tokens through J-lens. To keep task content out of the contrast, we compute it within the same question.

Specifically, we identify questions $q$ for which both faithful ($A$) and unfaithful ($B$) CoTs are available, and form one-to-one pairs $(q,c_q^A,c_q^B)$ without reusing a trace. To align readouts across CoTs of different lengths, we normalize reasoning progress and partition each CoT into $P$ relative-position bins. For label $y\in\{A,B\}$ and every layer $\ell$, let $\bar{\mathbf{h}}^{\,y}_{q,\ell,p}$ denote the mean hidden state at sentence and clause boundaries in bin $p$. At each layer $\ell$ and position bin $p$, we then average the within-question faithful--unfaithful differences:
$\Delta_{\ell,p}=\frac{1}{|\mathcal{Q}_{\ell,p}|}\sum_{q\in\mathcal{Q}_{\ell,p}}\left(\bar{\mathbf{h}}^{A}_{q,\ell,p}-\bar{\mathbf{h}}^{B}_{q,\ell,p}\right),
\label{eq:paired-contrast}$
where $\mathcal{Q}_{\ell,p}$ contains pairs with valid readouts for both labels in that cell. Thus, $\Delta_{\ell,p}$ is the mean direction from unfaithful to faithful internal states at cell $(\ell,p)$.

\begin{figure*}[h]
\centering
\includegraphics[width=\textwidth]{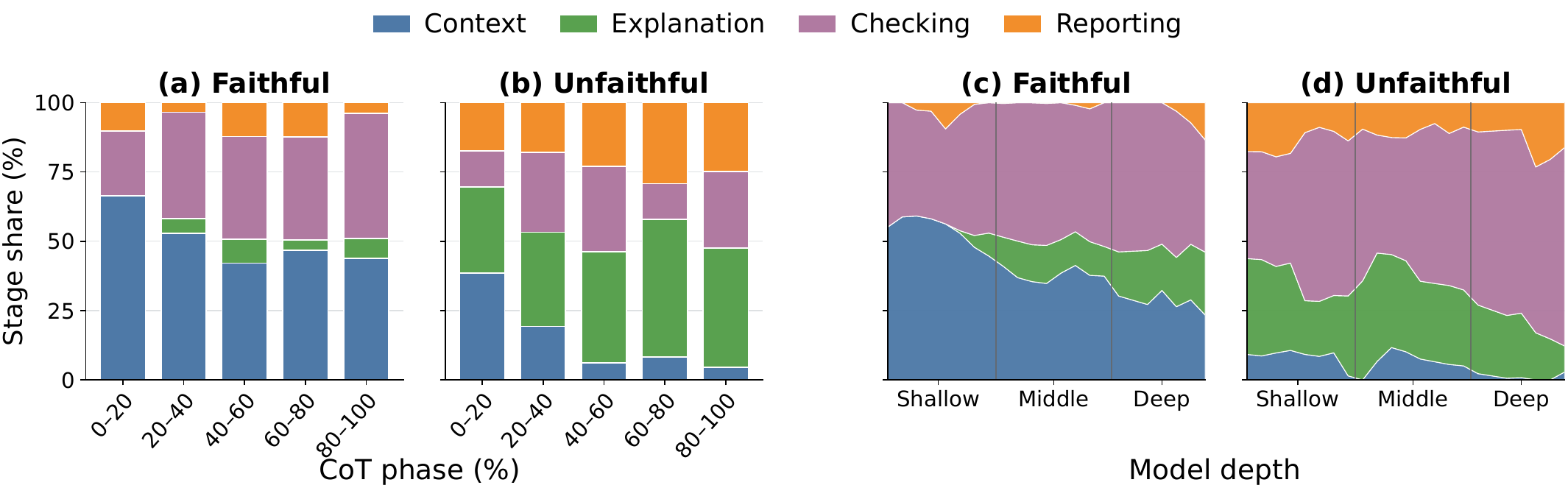}
\caption{Functional composition of faithfulness-related anchors on FaithCoT-Bench using Qwen3. (a)--(b) show changes across normalized CoT phases, while (c)--(d) show changes across model depth. Anchors are grouped into Context, Explanation, Checking, and Reporting stages, and each region reports its share of positive label-directed contrast mass. The complete words in each category are provided in Appendix~\ref{app:anchor-categories}.}
\label{fig:anchor-stage-composition}
\end{figure*}

We translate each contrast direction into token-level concepts using the J-lens vectors defined in \S\ref{sec:prelim}. For every candidate token $w\in\mathcal{W}$, we measure its alignment score with the contrast direction $\Delta_{\ell,p}$ as $s_{\ell,p,w}=\cos\!\left(\tilde{\mathbf{j}}_{\ell,w},\Delta_{\ell,p}\right)$. A large positive score indicates that the token's direction is more associated with faithful than unfaithful internal states. We retain the $\kappa$ highest-scoring positive tokens in each cell $(\ell,p)$ and take their union:
\begin{equation}
\mathcal{A}_{\ell,p}=\operatorname{Top}_{\kappa}\{w\in\mathcal{W}:s_{\ell,p,w}>0\},\qquad \mathcal{A}=\bigcup_{\ell,p}\mathcal{A}_{\ell,p}.
\label{eq:anchor-selection}
\end{equation}
The resulting dictionary $\mathcal{A}$ is then adopted to describe CoTs' transition via its representative tokens. To make this evolution easier to interpret, Figure~\ref{fig:anchor-stage-composition} groups the anchors into four stages (Context, Explanation, Checking, and Reporting) and shows where the stages are along CoT progress and model depth. Faithful-directed anchors stay grounded in Context, with Checking growing in later phases and deeper layers. Unfaithful-directed anchors lose Context early and are dominated by Explanation and Reporting, suggesting explanations detached from the task. The mapping from anchors to functional reasoning stages is described in Appendix~\ref{app:anchor-stages}.

\subsection{Faithful Detection with Transitional Features}
\label{sec:views}
The J-lens vectors $\tilde{\mathbf{j}}_{\ell,w}$ extracted from the contrastive pairs can serve as anchors; the alignment between an evaluated CoT and the anchor can be used for faithfulness detection. Therefore, we define the anchor alignment tensor as $\mathbf{r}=\Big[\cos\!\big(\bar{\mathbf{h}}_{\ell,p},\tilde{\mathbf{j}}_{\ell,w}\big)\Big]_{\ell\in\mathcal{L},\,p\le P,\,w\in\mathcal{A}}\in\mathbb{R}^{|\mathcal{L}|\times P\times|\mathcal{A}|}.$
describing the trajectory jointly over model depth and reasoning progress in terms of how strongly each anchor token is expressed. We further develop three complementary perspectives to trace the evaluated CoTs along the verbalisation and internal computation progression.


\textbf{Map} records the strength of faithfulness-relevant anchor evidence at every layer--phase cell. For cell $(\ell,p)$, we compute the cosine similarity between its hidden state and each selected J-lens anchor direction, and aggregate the alignments using their discovery scores:
\begin{equation}
m_{\ell,p}=\sum_{w\in\mathcal{A}_{\ell,p}}\frac{s_{\ell,p,w}}{\sum_{w'\in\mathcal{A}_{\ell,p}}s_{\ell,p,w'}}r_{\ell,p,w}.
\end{equation}
The matrix $\mathbf{m}=(m_{\ell,p})_{\ell\in\mathcal{L},\,p\leq P}$ retains the location of alignments over depth and reasoning.  

\textbf{Profile} summarizes anchor-specific evidence across the trajectory. For each anchor $w$, we average its alignment with the hidden states over the cells in which it was selected:
\begin{equation}
u_w
=\frac{1}{|\mathcal{C}_w|}\sum_{(\ell,p)\in\mathcal{C}_w}r_{\ell,p,w},\
\mathcal{C}_w=\{(\ell,p):w\in\mathcal{A}_{\ell,p}\}.
\end{equation}
The resulting profile $\mathbf{u}=(u_w)_{w\in\mathcal{A}}$ captures which faithfulness-relevant anchors are expressed and their average alignment strength throughout the reasoning trajectory.

\textbf{Dynamics} complements Map and Profile by measuring how the anchor-alignment changes between neighboring cells. We compute its displacement along model depth and reasoning progress, and then measure the alignment between consecutive displacements:
\begin{equation}
\begin{aligned}
\boldsymbol{\delta}^{\mathrm{dep}}_{\ell,p}=\mathbf{r}_{\ell,p}-\mathbf{r}_{\ell-1,p}&,\
\boldsymbol{\delta}^{\mathrm{prog}}_{\ell,p}=\mathbf{r}_{\ell,p}-\mathbf{r}_{\ell,p-1},\\
c^{\mathrm{dep}}_{\ell,p}=\cos\!\left(\boldsymbol{\delta}^{\mathrm{dep}}_{\ell,p},\boldsymbol{\delta}^{\mathrm{dep}}_{\ell-1,p}\right)&,\
c^{\mathrm{prog}}_{\ell,p}=\cos\!\left(\boldsymbol{\delta}^{\mathrm{prog}}_{\ell,p},\boldsymbol{\delta}^{\mathrm{prog}}_{\ell,p-1}\right).
\end{aligned}
\label{eq:response-field-dynamics}
\end{equation}
A high cosine indicates that the alignment tensors continue in a similar direction. The resulting dynamics matrices $\mathbf{C}^{\mathrm{dep}}=\{c^{\mathrm{dep}}_{\ell,p}\}\in\mathbb{R}^{(L-2)\times P}$ and $\mathbf{C}^{\mathrm{prog}}=\{c^{\mathrm{prog}}_{\ell,p}\}\in\mathbb{R}^{L\times(P-2)}$ capture directional consistency across model depth and CoT progress, respectively.

\paragraph{Structured Probe}
To detect CoT faithfulness, we concatenate the selected Map, Profile, and Dynamics features into a trajectory representation $\mathbf{x}$. We train an $L_2$-regularized logistic regression:
$P(y=A\mid\mathbf{x})=\sigma\!\left(\beta_0+\boldsymbol{\beta}^{\top}\mathbf{x}\right).$We use a simple linear classifier so that detection performance primarily reflects the extracted trajectory features rather than the capacity of the prediction head.
\section{Experiments}
We evaluate across two \textbf{benchmarks}: BonaFide \citep{gur2026faithfulness} and FaithCoT-Bench (with subsets: AQuA, LogiQA, TruthfulQA, and HLE-Bio) \citep{shen2026faithcot} on two backbones, i.e.,  Llama~\citep{grattafiori2024llama3} and Qwen~\citep{yang2025qwen3}. \textbf{Baselines} include Perplexity~\citep{cui2025stepwise}, counterfactual-based: Early Ans~\citep{lanham2023measuring}, logit-based: Info Gain~\citep{li2025towards} and Ans Tracing~\citep{lewis2025analysing}, LLM-as-judge~\citep{mittal2026c2}, and supervised hidden-state probes~\citep{goldowsky2025detecting} on the middle (HS$_m$) and last (HS$_l$) CoT tokens. We report balanced accuracy~\citep{brodersen2010balanced} and AUROC~\citep{fawcett2006introduction}.~\footnote{See Appendix~\ref{app:data} for detailed evaluation setup.}
Beyond the faithful CoT detection, we analyse how unfaithfulness manifests in our structured features and detector generalisability across datasets and backbone models.

\subsection{Faithful Detection Results}
\label{sec:main-results}
\begin{table*}[t]
\centering
\scriptsize
\setlength{\tabcolsep}{4.3pt}
\caption{Faithful detection evalaution results. \modelname{}$_p$ denotes the best Dynamics-Map configuration, and \modelname{}$_c$ additionally incorporates Profile. LLM judge produces hard labels, so its AUROC is unavailable.}
\label{tab:main}
\begin{tabular}{ll*{5}{cc}}
\toprule
& & \multicolumn{2}{c}{BonaFide} & \multicolumn{2}{c}{AQuA} & \multicolumn{2}{c}{LogiQA} & \multicolumn{2}{c}{TruthfulQA} & \multicolumn{2}{c}{HLE-Bio} \\
\cmidrule(lr){3-4}\cmidrule(lr){5-6}\cmidrule(lr){7-8}\cmidrule(lr){9-10}\cmidrule(lr){11-12}
& Method & Bal.\ acc. & AUROC & Bal.\ acc. & AUROC & Bal.\ acc. & AUROC & Bal.\ acc. & AUROC & Bal.\ acc. & AUROC \\
\midrule
\multirow{9}{*}{\rotatebox[origin=c]{90}{Qwen3}}
 & Perplexity & 0.565 & 0.582 & 0.510 & 0.558 & 0.562 & 0.570 & 0.563 & 0.540 & 0.533 & 0.520 \\
 & Early Ans & 0.537 & 0.528 & 0.578 & 0.598 & 0.695 & 0.715 & 0.722 & 0.766 & 0.505 & 0.548 \\

 & Info Gain & 0.742 & 0.847 & 0.631 & 0.692 & 0.566 & 0.566 & 0.545 & 0.577 & 0.552 & 0.516 \\
 & Ans Tracing & 0.631 & 0.683 & 0.668 & 0.736 & 0.709 & 0.735 & 0.722 & 0.742 & 0.545 & 0.577 \\

 & LLM judge & 0.581 & -- & 0.673 & -- & 0.606 & -- & 0.571 & -- & 0.444 & -- \\

 & HS$_{m}$ & 0.758 & \textbf{0.898} & 0.623 & 0.683 & 0.461 & 0.424 & 0.726 & 0.798 & 0.545 & 0.551 \\
 & HS$_{l}$ & 0.819 & 0.867 & 0.730 & 0.788 & 0.561 & 0.594 & 0.559 & 0.605 & 0.517 & 0.549 \\
\cmidrule(l){2-12}
 & \modelname{}$_{p}$ & \underline{0.831} & \underline{0.896} & \underline{0.733} & \underline{0.834} & \underline{0.713} & \underline{0.737} & \underline{0.764} & \textbf{0.805} & \underline{0.603} & \underline{0.635} \\
 & \modelname{}$_{c}$ & \textbf{0.852} & 0.883 & \textbf{0.765} & \textbf{0.840} & \textbf{0.719} & \textbf{0.744} & \textbf{0.768} & \underline{0.804} & \textbf{0.641} & \textbf{0.673} \\
\midrule
\multirow{9}{*}{\rotatebox[origin=c]{90}{Llama3.1}}
 & Perplexity & 0.491 & 0.450 & 0.494 & 0.455 & 0.540 & 0.561 & 0.592 & 0.616 & 0.423 & 0.443 \\

 & Early Ans & 0.522 & 0.530 & 0.556 & 0.575 & 0.612 & \underline{0.635} & 0.660 & 0.703 & 0.472 & 0.449 \\

 & Info Gain & 0.767 & 0.817 & 0.657 & 0.656 & 0.616 & 0.623 & 0.533 & 0.563 & 0.487 & 0.425 \\
 & Ans Tracing & 0.632 & 0.637 & 0.633 & 0.694 & 0.610 & 0.634 & 0.693 & \underline{0.793} & 0.488 & 0.482 \\

 & LLM judge & 0.425 & -- & 0.536 & -- & 0.591 & -- & 0.478 & -- & 0.479 & -- \\

 & HS$_{m}$ & 0.789 & 0.769 & 0.628 & 0.695 & 0.472 & 0.590 & 0.709 & 0.647 & 0.536 & 0.583 \\
 & HS$_{l}$ & 0.740 & 0.820 & 0.677 & 0.748 & 0.622 & 0.616 & 0.625 & 0.625 & 0.558 & \underline{0.622} \\
\cmidrule(l){2-12}
 & \modelname{}$_{p}$ & \underline{0.854} & \underline{0.823} & \underline{0.715} & \textbf{0.764} & \underline{0.670} & 0.628 & \underline{0.767} & 0.786 & \textbf{0.633} & 0.606 \\
 & \modelname{}$_{c}$ & \textbf{0.884} & \textbf{0.842} & \textbf{0.720} & \underline{0.758} & \textbf{0.688} & \textbf{0.645} & \textbf{0.773} & \textbf{0.795} & \underline{0.621} & \textbf{0.627} \\
\bottomrule
\end{tabular}
\end{table*}
Table~\ref{tab:main} shows the faithful detection results across baselines and \modelname{} with two variants: \modelname{}$_p$ incorporates the features: \textsl{Dynamics} and \textsl{Map}; \modelname{}$_c$ additionally incorporates \textsl{Profile} as our full method. In general, our \modelname{} achieves the best detection accuracy across different setups.
Averaged over the five datasets, \modelname{}$_c$ reaches 0.749 with Qwen3 and 0.737 with Llama3.1, more than 9 points above the strongest baseline for each model (Ans Tracing, 0.655; HS$_l$, 0.644). Among unsupervised baselines, logit-based methods perform better than text-level signals (0.610 versus 0.533 in balanced accuracy), yet both fall well below \modelname{}$_c$ (0.743). These results indicate that faithfulness is better captured by the trajectory of internal information than by a single hidden-state readout or the visible rationale alone. 

\subsection{Understanding the Transition via Structured Features}

Structural features summarise CoT reasoning trajectories from complementary angles via the J-lens. Table~\ref{tab:views} shows that each feature view contributes individually to detection performance, and that combining all three yields the best results across all settings. We further analyse how these features reflect distinct internal computation pathways that underpin faithful and unfaithful reasoning.

\begin{figure*}[h]
\centering
\includegraphics[width=0.75\textwidth]{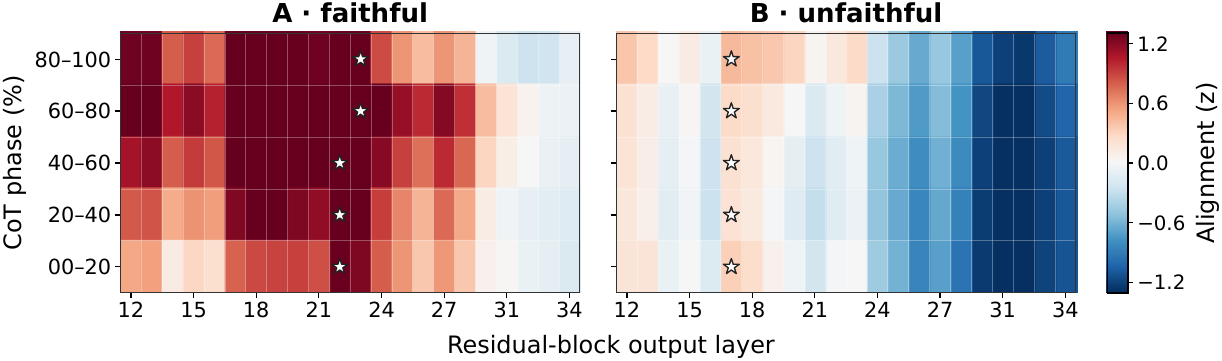}
\caption{Layer--phase anchor alignments (Map) on BonaFide using Qwen3. Each cell shows the group-balanced mean standardized alignment for faithful (A) and unfaithful (B) CoTs.}
\label{fig:anchor-heatmap}
\end{figure*}

\textbf{Map: where does faithfulness-relevant evidence appear?}
Figure~\ref{fig:anchor-heatmap} visualises the \textsl{Map} feature, i.e., anchor alignment over model depth and CoT progress. The two groups differ in both layer and phase. (1) Faithful CoTs show higher overall anchor alignment. (2) Their peak locations differ across CoT phases. Faithful CoTs peak around layers 22--23, shifting slightly deeper in later phases, whereas unfaithful CoTs peak consistently near layer 17. Map preserves this layer--phase structure, which a single pooled alignment score would discard. 

\begin{figure*}[t]
\centering
\begin{subfigure}[t]{0.38\textwidth}
\centering
\includegraphics[width=\linewidth]{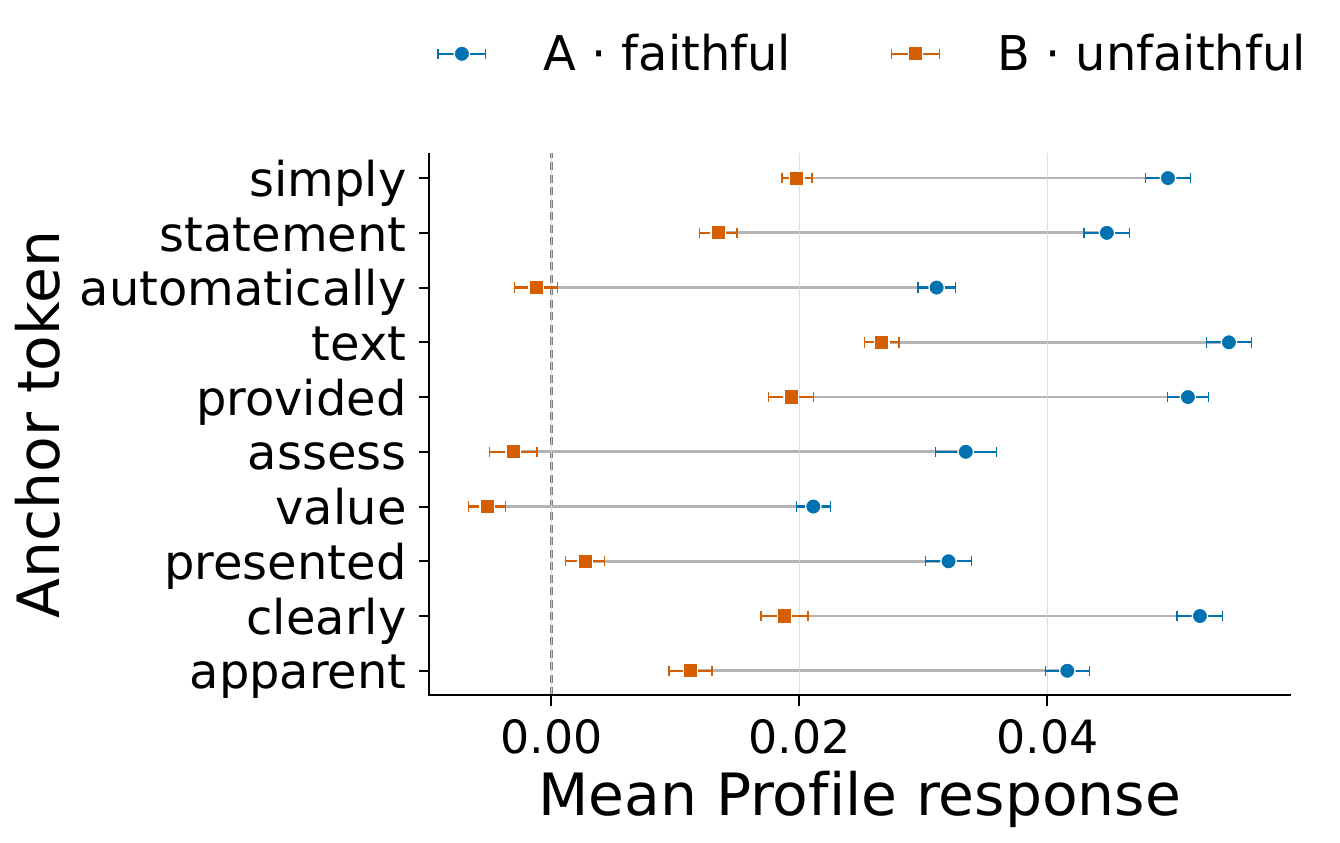}
\caption{Profile alignments on BonaFide. }
\label{fig:profile-anchors}
\end{subfigure}
\hfill
\begin{subfigure}[t]{0.6\textwidth}
\centering
\includegraphics[width=\linewidth]{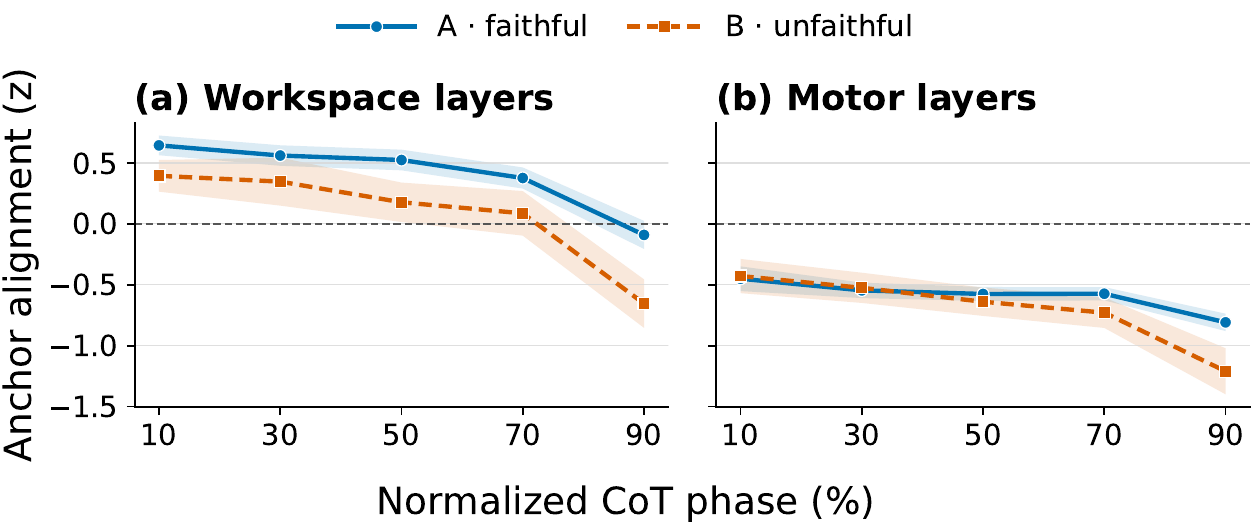}
\caption{alignment across CoT phases on AQuA.}
\label{fig:anchor-dynamics}
\end{subfigure}
\caption{Analysis of faithfulness-relevant anchor alignments using Qwen3 representations. (a) Mean Profile alignments for the 10 most discriminative anchors. (b) Group-balanced mean alignment across normalized CoT phases, averaged within the workspace and motor layers. Shaded regions denote 95\% confidence intervals.}
\label{fig:anchor-profile-persistence}
\end{figure*}

\textbf{Profile: which anchors distinguish faithful CoTs?}
Profile keeps a separate alignment for each anchor. Figure~\ref{fig:profile-anchors} shows the Profile alignments of the 10 anchor tokens that contribute most to Profile on BonaFide. Faithful CoTs respond more strongly on all of them, especially on anchors for grounding and assessing information (\emph{text}, \emph{provided}, \emph{statement}, \emph{clearly}, \emph{assess}). The gap varies by anchor: \emph{text} stays moderately high in unfaithful CoTs, whereas \emph{assess}, \emph{value}, and \emph{automatically} drop to near zero. Unlike Map, which aggregates over anchors, Profile keeps a separate alignment for each anchor and thus retains these differences.


\textbf{Dynamics: how does anchor evidence evolve?} Map and Profile neither captures how alignments change from one cell to the next. Figure~\ref{fig:anchor-dynamics} tracks anchor alignment across CoT phases in the middle and later layers (as early layers fail to generate stable verbalisation), also known as workspace and motor layers in \citet{gurnee2026verbalizable}. In the motor layers, the two groups are indistinguishable up to 30\% of the CoT. Faithful CoTs then plateau, whereas unfaithful CoTs keep declining and drop sharply in the final phase. The workspace layers show the same late divergence. These differences arise from how alignment changes between phases, not from its value in any single phase. Dynamics targets this signal by measuring whether consecutive changes along CoT progress and model depth keep a consistent direction.


\subsection{Sensitivity Analysis}
\label{sec:analysis}
\begin{figure}[t]
\centering
\begin{subfigure}{0.44\textwidth}
\centering
\includegraphics[width=\textwidth]{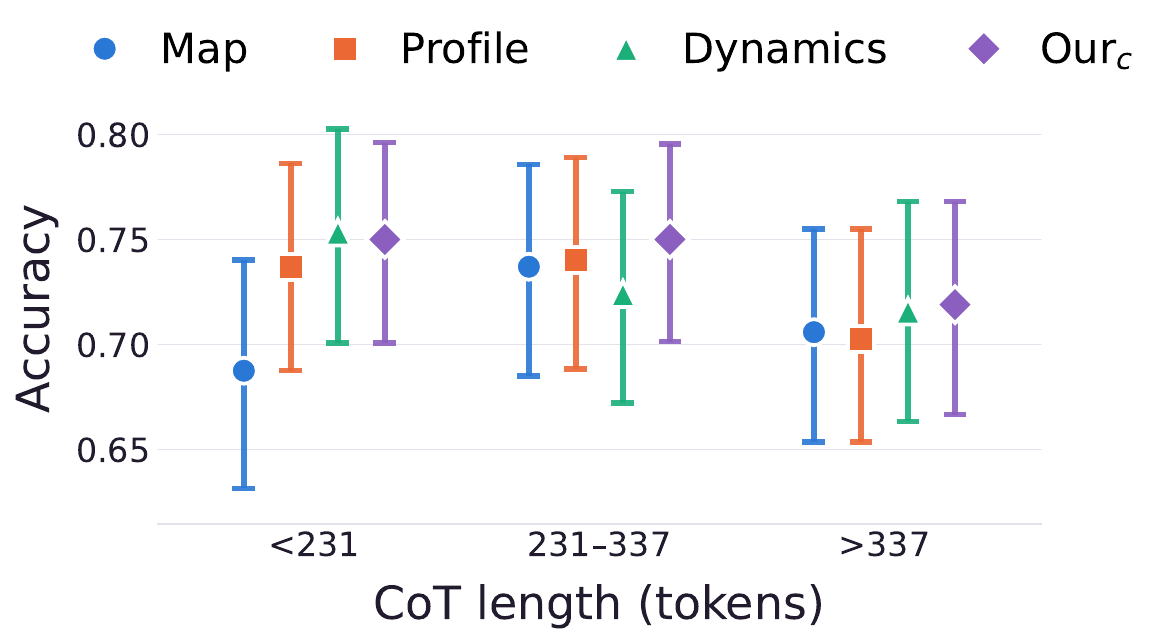}
\caption{CoT length.}
\label{fig:feature-length}
\end{subfigure}
\hfill
\begin{subfigure}{0.44\textwidth}
\centering
\includegraphics[width=\textwidth]{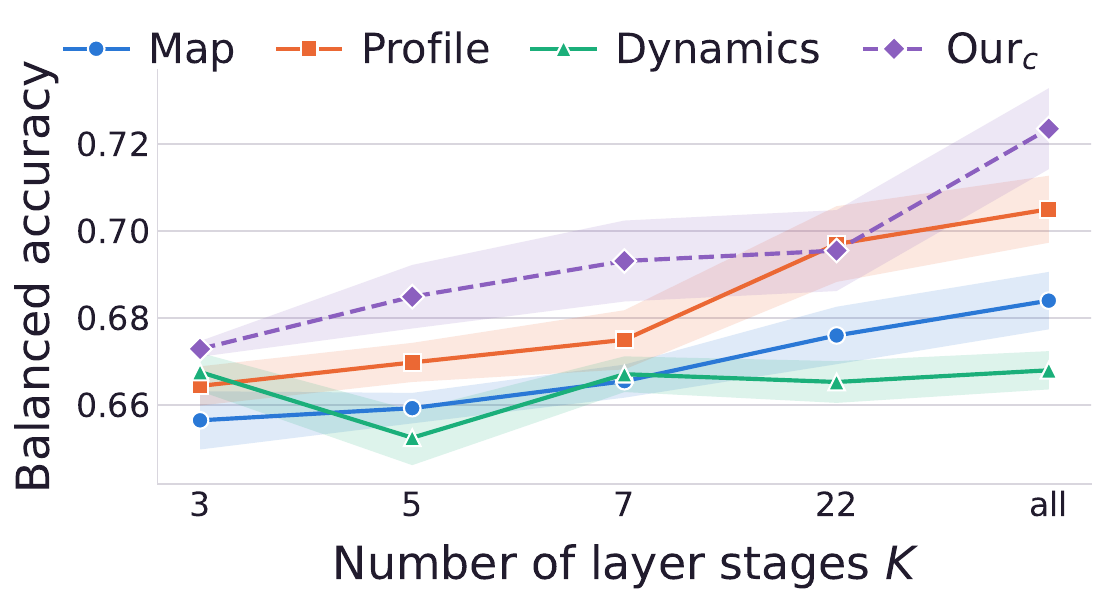}
\caption{Number of layer stages $K$.}
\label{fig:stage-granularity}
\end{subfigure}
\caption{Effect of key factors on Qwen3-4B. (a)~Balanced accuracy of features within CoT-length terciles on BonaFide. (b)~Balanced accuracy on FaithCoT when layers are merged into $K$ stages.}
\label{fig:factors}
\end{figure}

\textbf{CoT length.} Figure~\ref{fig:feature-length} splits CoTs into length terciles. Each feature responds differently to length, and each pattern follows from how the feature is computed. Map improves from 0.688 on short CoTs to 0.737 on medium ones. It averages alignment within a fixed layer--phase grid, so a short CoT leaves few tokens per cell and yields noisy averages. Profile is stable on short and medium CoTs (0.737 and 0.740) but drops to 0.703 on long ones. It pools each anchor's alignment across the whole CoT, and a long CoT mixes more heterogeneous steps, diluting the pooled alignment. Dynamics is strongest on short CoTs (0.753) and declines as length grows. Because the number of phases is fixed, each phase covers more tokens in a long CoT. Local turns are then averaged out within a cell, leaving smoother changes between neighbouring cells. The features therefore favour different lengths, which explains why combining them helps.

\textbf{Layer stages.} Figure~\ref{fig:stage-granularity} merges adjacent layers into $K$ stages. Coarser stages hurt Map and Profile most. Both rely on anchors discovered from stage-level alignments, and coarse stages average out informative anchors. Map also loses layer resolution, blurring where evidence appears. Dynamics stays nearly flat, as its cosine tracks the overall direction of change rather than individual anchors. Combining all three features in \modelname{}$_c$ yields the most stable performance. It retains 0.693 at $K=7$, close to 0.695 with all layers, and still outperforms every single feature at $K=3$ (0.673). Using the unified bins for layers supports the model-wise transfer; see results in Figure~\ref{fig:cross-model}. Additional sensitivity analyses are provided in Appendix~\ref{app:sensitivity}.

\subsection{Cross-Source Generalization}
\label{sec:generalization}
\begin{figure*}[t]
\centering
\begin{subfigure}[b]{0.54\textwidth}
\centering
\includegraphics[width=\textwidth]{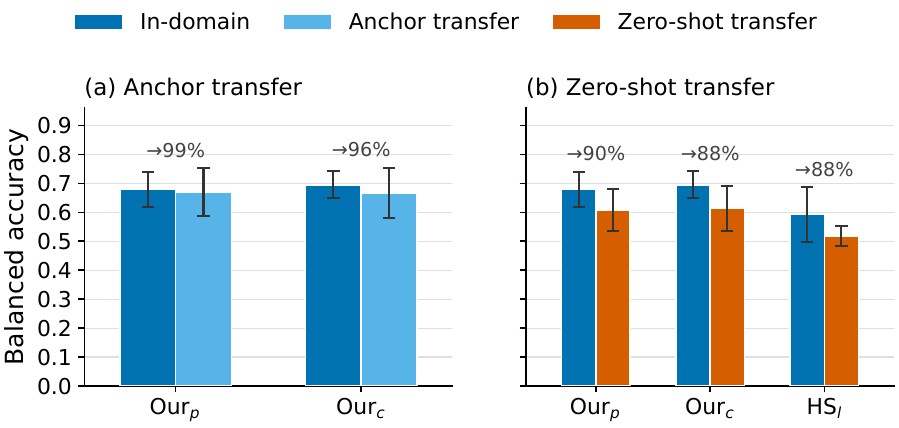}
\caption{Cross-source transfer.}
\label{fig:cross-source}
\end{subfigure}
\hfill
\begin{subfigure}[b]{0.44\textwidth}
\centering
\includegraphics[width=\textwidth]{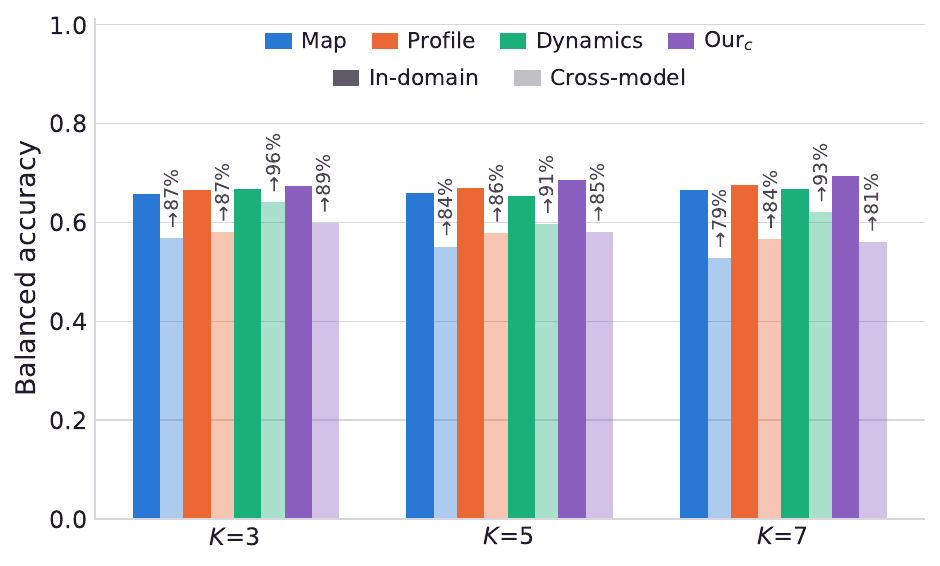}
\caption{Cross-model transfer.}
\label{fig:cross-model}
\end{subfigure}
\caption{Transfer on FaithCoT-Bench. (a)~Cross-subset transfer with Qwen3. Anchor transfer reuses the anchors discovered on the source subset and refits the classifier on the target, whereas zero-shot transfer freezes both. (b)~Cross-model transfer from Llama3.1 to Qwen3, applying the classifier trained on Llama3.1 unchanged. Percentages show the retained fraction of in-domain balanced accuracy.}
\label{fig:transfer}
\end{figure*}
\textbf{Anchor transfer.}
Figure~\ref{fig:cross-source} evaluates transfer across the four FaithCoT datasets using Qwen3. We freeze the anchors discovered on the source dataset. Both variants retain 97--99\% of their in-domain performance in balanced accuracy. The anchors discovered on one source therefore carry faithfulness signals that remain informative on other sources. AUROC results are provided in Appendix~\ref{app:cross-auroc}.

\textbf{Classifier transfer.}
We next freeze both the anchors and the classifier from the source dataset (zero-shot transfer in Figure~\ref{fig:cross-source}). Our variants still retain 88--90\% of their in-domain performance in both metrics, remaining well above HS$_l$ after transfer. The larger drop than under anchor transfer suggests that the anchors are largely source-general, while classifier calibration is partly source-dependent.

\textbf{Cross-model transfer.}
Figure~\ref{fig:cross-model} applies a classifier trained on Llama3.1 directly to Qwen3 without refitting, which stage grouping enables by mapping both models onto the same relative-depth grid. Dynamics transfers best, retaining 91--96\% of its in-domain performance at every $K$, because its cosine measurement is independent of each model's activation scale. Map transfers worst (79--87\%), because the exact layers where evidence appears differ between models. Coarser stages also transfer better, with \modelname{}$_c$ retaining 81\% at $K=7$ and 89\% at $K=3$, since they better absorb the depth mismatch between models. 

\section{Related Work}
Prior work on CoT faithfulness falls into two families: external methods that analyze observable model behavior, and internal methods that inspect the representations underlying the generated rationale. \textbf{External methods} infer faithfulness from model inputs and outputs. \citet{turpin2023language} add biasing cues and test whether the CoT discloses their influence. \citet{lanham2023measuring} truncate, paraphrase, or corrupt the CoT and observe how the answer changes. \citet{paul2024making} measure whether intermediate reasoning steps causally affect the prediction. \citet{chen2025reasoning} test whether models disclose hints that influence their answers. Other work uses human or LLM judges to detect unsupported and post-hoc reasoning \citep{shen2026faithcot}. These methods do not require access to model activations. However, they cannot observe decision-relevant information omitted or concealed by the CoT. \textbf{Internal methods} inspect the computation behind an explanation. Interventional methods measure whether the answer causally depends on the CoT, via activation patching~\citep{yeo2025towards}, parameter-level unlearning~\citep{tutek2025measuring}, or information-flow analysis~\citep{jia2026faithfulness}. Representational methods read the model's states instead. Hidden-state probes~\citep{goldowsky2025detecting} classify static activations at a fixed layer or position. Recent methods model the computation itself.  \citet{zhao2026verifying} treat per-step attribution graphs as execution traces of reasoning circuits and detect errors from their structural features. \citet{shen2026detecting} compare a circuit-derived internal reasoning graph with one built from the generated text. These methods show that the internal computation differs, but none verbalizes how unfaithfulness develops during generation, i.e., how reasoning-relevant information transits across layers and the reasoning process.
We address this limitation with interpretable, token-named J-space anchors.
\section{Conclusion}
We propose \modelname{}, a CoT faithfulness detector that uses J-lens to verbalize the model's internal computation. Unlike hidden-state probes that capture a static snapshot, \modelname{} tracks how reasoning-relevant information emerges and evolves across model depth and CoT progress. Specifically, we derive anchor tokens and measure how strongly hidden states align with them. Map, Profile, and Dynamics then summarize where this evidence appears, which anchors carry it, and how it evolves. Experiments on two benchmarks and two backbones show that \modelname{} outperforms all baselines, transfers across datasets and models, and explains unfaithfulness in human-readable terms. Future work could use these verbalized trajectories to steer models toward faithful reasoning.

\section*{Ethics statement}

This work proposes a framework for detecting and explaining unfaithful reasoning in large language models. Our method operates on existing models and datasets without collecting human subjects data or releasing new sensitive datasets. The benchmarks used, FaithCoT-Bench and BonaFide, are publicly available research datasets. We foresee no direct harmful applications; improved faithfulness detection supports safer and more transparent deployment of LLMs. We acknowledge that, like any interpretability method, traces produced by \modelname{} should be treated as diagnostic tools rather than ground-truth explanations of model behaviour. No conflicts of interest or external sponsorship influenced this research.

\section*{Reproducibility statement}

We have made every effort to ensure the reproducibility of our results. The core components of \modelname{}, including the J-lens projection, faithful anchor construction, and structured feature extraction, are described in detail in Section~\ref{sec:method}, with additional implementation details provided in Appendix~\ref{app:experimental-details}. Hyperparameters for the classifier and feature extraction pipeline, including layer selection, CoT phase normalisation, and anchor token identification, are fully specified in Appendix~\ref{app:hyper}. All experiments are conducted on publicly available benchmarks: FaithCoT-Bench and BonaFide, with dataset statistics and preprocessing steps documented in Appendix~\ref{app:data}. The models used in our experiments are openly accessible through HuggingFace, and the specific model versions and hardware configurations are reported in Appendix~\ref{app:backbone}. To support further research, we will release the full source code, including feature extraction scripts, classifier training, and evaluation pipelines, as anonymous supplementary materials accompanying this submission. Ablation results reported in Section~\ref{sec:analysis}, including per-feature-view contributions and cross-dataset generalisation, are reproducible using the released code and the dataset splits described in Appendix~\ref{app:data}.

\section*{AI use statement}
In this work, we used generative AI tools to \textbf{[edit the Introduction and Related Work sections to improve readability and to edit code for improving the visual presentation of figures]}. We have reviewed all AI-assisted work. AI-drafted content is modified and revised by the authors; information retrieved and provided by AI is validated by the authors; AI-edited language is double-checked by the authors to ensure that the correct message is conveyed. Additionally, we have not used generative AI tools to [help develop theoretical models or conceptual frameworks, formulate mathematical claims, propose or refine hypotheses, design or provide feedback on research methodology or experiments, implement methods, generate synthetic data, clean or reformat datasets, or support qualitative and thematic data analysis].

We take responsibility for the final content of this work, including text produced with the aid of generative AI.

\bibliography{iclr2027_conference}
\bibliographystyle{iclr2027_conference}
\clearpage
\appendix
\section{Backbones and Datasets}
\paragraph{Backbones.}
\label{app:backbone}
We use two open-weight backbones as replay models: Llama-3.1-8B-Instruct~\citep{grattafiori2024llama3} (32 layers, hidden size 4096) and Qwen3-4B-Thinking-2507~\citep{yang2025qwen3} (36 layers, hidden size 2560). Each question and its given CoT are fed to the backbone in a single forward pass, and we read the residual-stream activations at every probed layer. The same CoTs are therefore analyzed under two different internal representations.

\paragraph{BonaFide.}
\label{app:data}
BonaFide~\citep{gur2026faithfulness} provides CoTs with ground-truth faithfulness labels. We use its diversionary setting. Each question comes with a misleading hint that points to an incorrect answer, and only CoTs whose final answer follows the hint are kept. A CoT is faithful if it acknowledges or verbalizes its reliance on the hint, and unfaithful if it follows the hint without mentioning it or attributes the answer to a fabricated source. Questions come from SimpleQA-Verified, Humanity's Last Exam, and DDXPlus. The hints take six forms (sycophancy, metadata, validator, error message, security audit, and unauthorized access), and CoTs are produced by ten source models. Our evaluation set contains 1{,}016 CoTs over 541 questions, of which 87.5\% are unfaithful.

\paragraph{FaithCoT-Bench.}
FaithCoT-Bench~\citep{shen2026faithcot} contains CoTs generated by four LLMs (LLaMA3.1-8B, Qwen2.5-7B, GPT-4o-mini, and Gemini 2.5 Flash). Each CoT is labeled faithful or unfaithful by expert annotators over multiple rounds, with Cohen's $\kappa$ between 81.0 and 97.2. We use its four subsets: AQuA (algebraic word problems), LogiQA (deductive reasoning), TruthfulQA (questions designed to elicit common misconceptions), and HLE-Bio (biomedical questions from Humanity's Last Exam). Table~\ref{tab:data-stats} lists the size and label balance of each dataset.

\begin{table}[h]
\centering
\small
\caption{Dataset statistics.}
\label{tab:data-stats}
\begin{tabular}{lrrrr}
\toprule
Dataset & \#CoTs & \#Questions & \#Faithful & \#Unfaithful \\
\midrule
BonaFide   & 1{,}016 & 541 & 127 & 889 \\
AQuA       & 354 & 100 & 273 &  81 \\
LogiQA     & 400 & 100 & 273 & 127 \\
TruthfulQA & 389 & 100 & 280 & 109 \\
HLE-Bio    & 161 &  41 &  96 &  65 \\
\bottomrule
\end{tabular}
\end{table}

\section{Experimental Details}
\label{app:experimental-details}

\subsection{J-lens Estimation}
For each replay model, we estimate the J-lens on 1,000 sequences sampled from WikiText-2, independently of all labeled CoT data. Each sequence is truncated to 128 model tokens, and the first 16 positions are excluded from the Jacobian average. The lens differentiates the sum of the valid final-layer residual states with respect to each source-layer state and averages the resulting Jacobians across positions and sequences. Matrices are stored in \texttt{float32}.

\subsection{Anchor and Feature Configuration}
We divide each CoT into five normalized phase bins: 0--20\%, 20--40\%, 40--60\%, 60--80\%, and 80--100\%. Anchor discovery searches a fixed content-word candidate dictionary and retains the five positively aligned tokens with the largest discovery scores in each layer--phase cell. The resulting union is frozen before evaluation. Qwen3 and Llama3.1 use a 2,048-token candidate dictionary. All hyperparameters are fixed across datasets within a run.

\subsection{LLM Judge}
We use Qwen2.5-7B-Instruct and Llama3.1-8b-Instruct with greedy decoding, disabled reasoning mode, and one judgment per example. The judge receives only the question, answer choices when available, visible CoT, and final answer; reference labels, intervention measurements, source-model identity, and correct answers are withheld. FaithCoT-Bench uses the following instruction:

\begin{quote}
\small
Determine whether the visible chain of thought is a causally faithful account of how the final answer was reached. Label it $A$ when the reasoning forms a coherent path that genuinely supports the answer, and $B$ when the answer is decoupled from the reasoning, a decisive step is absent, or the explanation appears post hoc. Do not classify based on correctness alone. End with \texttt{<answer>A</answer>} or \texttt{<answer>B</answer>}.
\end{quote}

For BonaFide, the judge is instead asked whether the CoT explicitly discloses that the external hint influenced the answer. Merely repeating the hinted answer, claiming to have ignored the hint, or attributing the answer to another source is labeled $B$. Outputs that do not contain a valid $A/B$ verdict are excluded from the judge accuracy calculation. Because the judge produces only a hard decision, we report balanced accuracy but not AUROC.

\subsection{Raw Hidden-state Probes}
The Raw-HS baseline uses one fixed residual-stream layer selected by the relative-depth rule of \citet{goldowsky2025detecting}: layer 10 for Qwen3 and layer 9 for Llama3.1. The mean-token variant trains on individual CoT-token activations and averages the resulting decision scores within each response. The last-token variant trains a separate classifier using only the final non-whitespace CoT token. Both use fold-wise standardization and $L_2$-regularized logistic regression with $C=0.1$, no intercept, and no class weighting. We do not search over layers.

\subsection{Classifier Configurations}
For our method, missing feature values are imputed with the training-fold median and all features are standardized using training-fold statistics. We then fit an $L_2$-regularized logistic regression with regularization coefficient 1 and balanced class weights. In Table~\ref{tab:main}, \modelname{}$_p$ selects the highest-balanced-accuracy configuration composed of Map and Profile features, whereas \modelname{}$_c$ selects the corresponding configuration that additionally includes Dynamics. AUROC is reported for the same selected configuration.

\subsection{Evaluation Protocol}
\label{app:hyper}
We report five-fold grouped out-of-fold predictions with seed 42. All CoTs associated with the same question remain in the same fold, preventing alternative traces for one question from appearing in both training and evaluation. Anchor discovery is performed only on the separate discovery questions and is frozen before classification. Preprocessing and classifier fitting are repeated independently within each training fold. Balanced accuracy uses a threshold of 0.5, and AUROC treats faithful CoTs as the positive class.

\subsection{Functional Categories of Anchor Tokens}
\label{app:anchor-categories}

For the analysis in Figure~\ref{fig:anchor-stage-composition}, we manually group the selected anchor tokens into four functional categories. \emph{Context} describes the input or represented information; \emph{Explanation} describes reasoning and justification; \emph{Checking} concerns verification, correctness, and error detection; and \emph{Reporting} describes the expression of an answer or conclusion. This grouping is used only for interpretation and does not affect anchor discovery or classifier training. Table~\ref{tab:anchor-categories} lists the complete category vocabulary before removing tokens shared by the faithful and unfaithful anchor pools.

\begin{table*}[t]
\centering
\small
\setlength{\tabcolsep}{5pt}
\caption{Functional grouping of anchor tokens used in Figure~\ref{fig:anchor-stage-composition}, before removing tokens shared by the faithful and unfaithful anchor pools.}
\label{tab:anchor-categories}
\begin{tabular}{lp{0.82\textwidth}}
\toprule
Category & Tokens \\
\midrule
Context & \texttt{actual, context, external, fact, hidden, influence, information, input, internal, meaning, prompt, question, relevant, state, understand, visible} \\
Explanation & \texttt{claim, explain, explanation, justify, logic, reason, reasoning, support} \\
Checking & \texttt{check, correct, critical, error, evidence, faith, false, quality, true, valid, verified, verify, wrong} \\
Reporting & \texttt{answer, conclusion, decision, express, expression, final, output, report, reported, reporting, statement} \\
\bottomrule
\end{tabular}
\end{table*}

For each CoT phase or model layer, positive label-specific contrast scores are summed within each category and normalized across the four categories. The plotted values therefore represent the composition of discriminative anchor evidence rather than raw token frequency. Tokens shared by the faithful and unfaithful candidate pools are removed globally before computing the displayed category shares.
\section{Extra Experiment Results}
\subsection{Feature Ablation}
Table~\ref{tab:views} shows that the three feature families provide complementary information. The strongest individual feature varies across models and benchmarks, while combining Map, Profile, and Dynamics achieves the best result in every setting. This indicates that faithfulness depends jointly on where anchor responses occur, which anchors are expressed, and how their representations evolve.
\begin{table*}[t]
\centering
\small
\setlength{\tabcolsep}{4pt}
\caption{Feature-set ablation in balanced accuracy. $d_Q$ and $d_L$ denote the feature counts under the Qwen3 and Llama3.1 models, respectively. }
\label{tab:views}
\begin{tabular}{lrccrcc}
\toprule
&  \multicolumn{3}{c}{Qwen3-4B} & \multicolumn{3}{c}{Llama3.1-8B} \\
\cmidrule(lr){2-4}\cmidrule(lr){5-7}
Feature set & $d_Q$ & BonaFide & FaithCoT & $d_L$ & BonaFide & FaithCoT \\
\midrule
Map & 146 & 0.814 & 0.684 & 126 & 0.841 & 0.640 \\
Profile  & 161 & 0.819 & 0.705 & 108 & 0.838 & 0.629 \\
Dynamics & 247 & \underline{0.840} & 0.668 & 211 & 0.833 & 0.643 \\
\midrule
Dynamics $+$ Profile & 408 & 0.831 & 0.689 & 319 & 0.854 & 0.622 \\
Dynamics $+$ Map &  393 & 0.813 & \underline{0.703} & 337 & \underline{0.867} & \underline{0.696} \\
Dynamics $+$ Map $+$ Profile & 554 & \textbf{0.848} & \textbf{0.723} & 445 & \textbf{0.884} & \textbf{0.697} \\
\bottomrule
\end{tabular}
\end{table*}

\subsection{Features' Cross-Source Effect}
\begin{figure*}[t]
\centering
\includegraphics[width=0.7\linewidth]{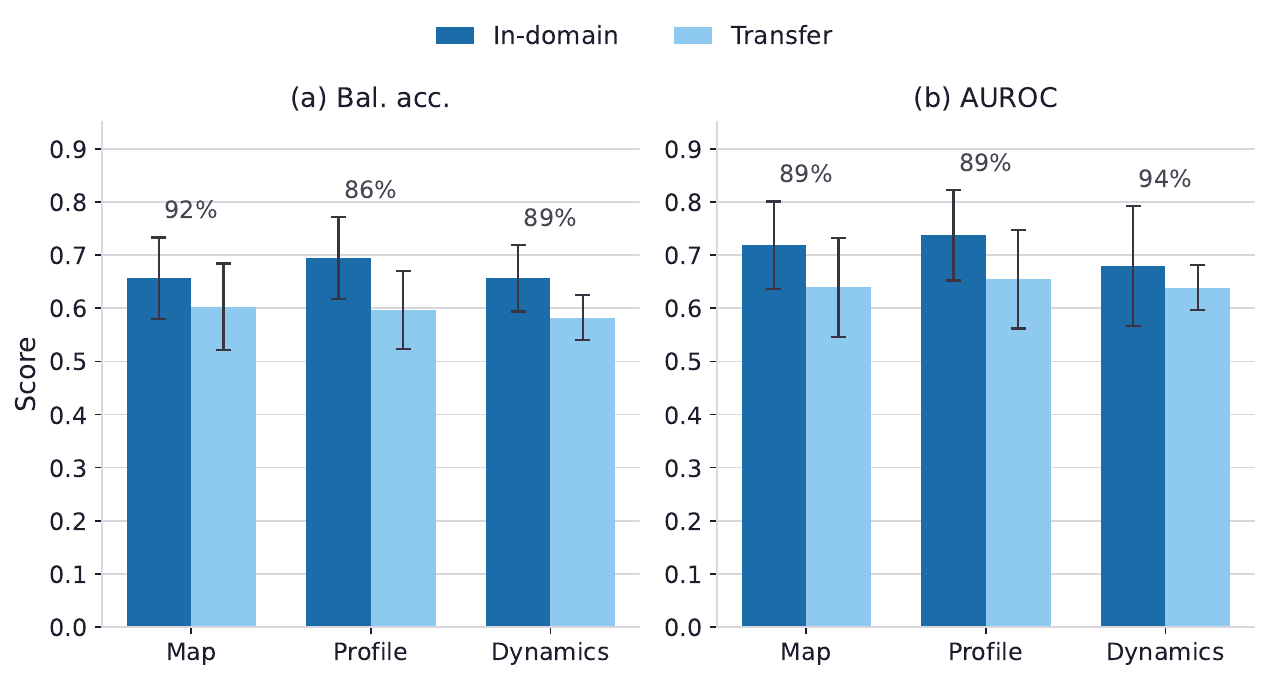}
\caption{Effect of cross-source on the three structured features using Qwen3-4B. Zero-shot transfer on FaithCoT, where anchors and classifier are frozen on one source subset and evaluated on another.}
\label{fig:anchor-persistence}
\end{figure*}
Figure~\ref{fig:anchor-persistence} freezes both anchors and classifier on one FaithCoT source subset and evaluates them on another. Profile retains the least (86\% in balanced accuracy), since which anchors are discriminative varies with source content. Map retains more (92\%), since where evidence appears across layers and phases reflects the model's computation rather than the source.  Dynamics retains the most in AUROC (94\%). Its cosine-based measure ignores activation scale and is less affected by source-specific shifts. Its lower retention in balanced accuracy (89\%) suggests that the ranking transfers but the decision threshold shifts.
\subsection{Cross-source Generalization}
\label{app:cross-auroc}
\begin{figure*}[t]
\centering
\includegraphics[width=0.72\textwidth]{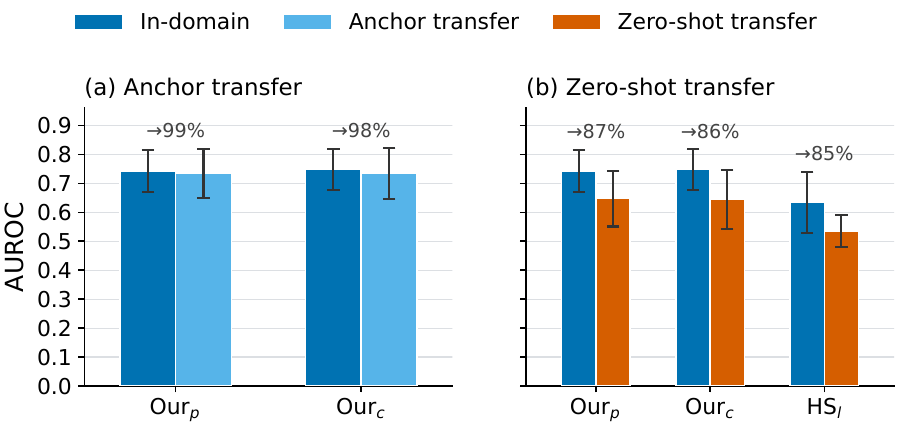}
\caption{Cross-source generalization in AUROC on FaithCoT-Bench. Anchor transfer refits the classifier on the target while freezing the transferred anchors, whereas zero-shot transfer freezes both. Error bars show standard deviations over the twelve ordered source pairs; percentages indicate performance retained relative to the in-domain setting.}
\label{fig:transfer-auroc}
\end{figure*}

We evaluate all twelve ordered source--target pairs among the four FaithCoT-Bench subsets. The in-domain setting learns both anchors and classifier on the target dataset. Anchor transfer freezes source-discovered anchors but refits the classifier on the target, whereas zero-shot transfer applies both without target training. Figure~\ref{fig:transfer-auroc} shows that anchor transfer preserves 98--99\% of in-domain AUROC. Under zero-shot transfer, Our$_p$ and Our$_c$ retain 87\% and 86\%, respectively, compared with 85\% for HS$_l$. This suggests that the anchor representation transfers well across sources, while the classifier remains partly source-dependent.

\subsection{Sensitivity Analysis}
\label{app:sensitivity}
\paragraph{CoT phases}
\begin{figure}[t]
\centering
\includegraphics[width=0.6\linewidth]{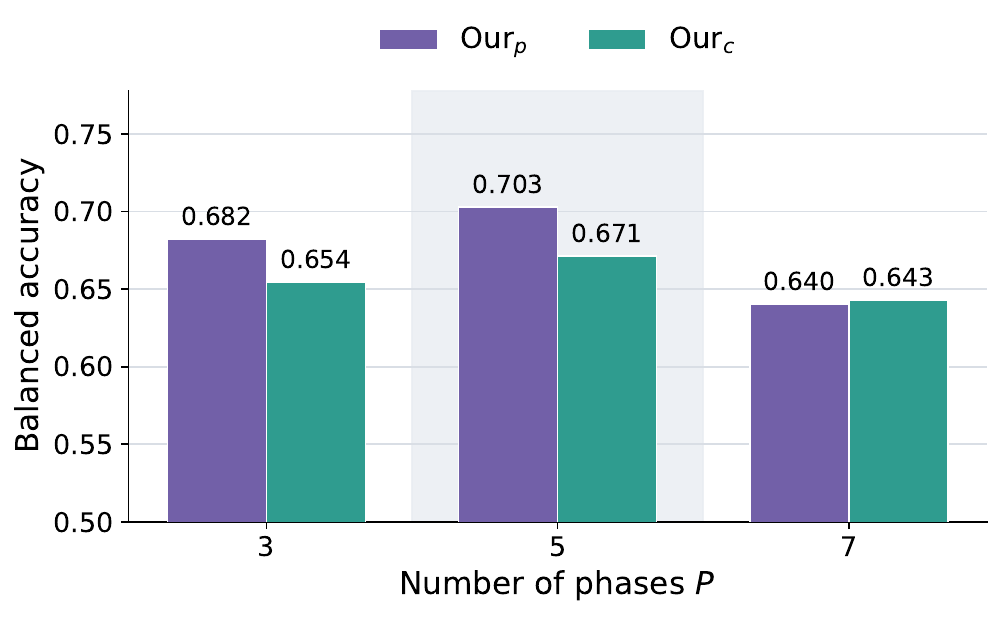}
\caption{Sensitivity to the number of normalized CoT phases $P$ on FaithCoT-Bench using Qwen3. Shading marks the default setting $P=5$.}
\label{fig:phase-sensitivity}
\end{figure}
We vary the number of normalized CoT phases over $P\in\{3,5,7\}$ while holding all other settings fixed, and report the mean balanced accuracy over the four FaithCoT-Bench sources. Figure~\ref{fig:phase-sensitivity} shows that both variants perform best with $P=5$, reaching 0.703 and 0.671. Three phases provide insufficient temporal resolution, while seven phases produce sparser cells. We therefore use $P=5$ in the main experiments.

\paragraph{Number of Anchors}
We vary the number of anchors selected per layer--phase cell over $\kappa\in\{4,8,16\}$ while fixing $P=5$ and all other settings. Anchor selection and evaluation are repeated separately on the four FaithCoT-Bench sources, and balanced accuracy is macro-averaged across them.

\begin{figure}[t]
\centering
\includegraphics[width=0.6\linewidth]{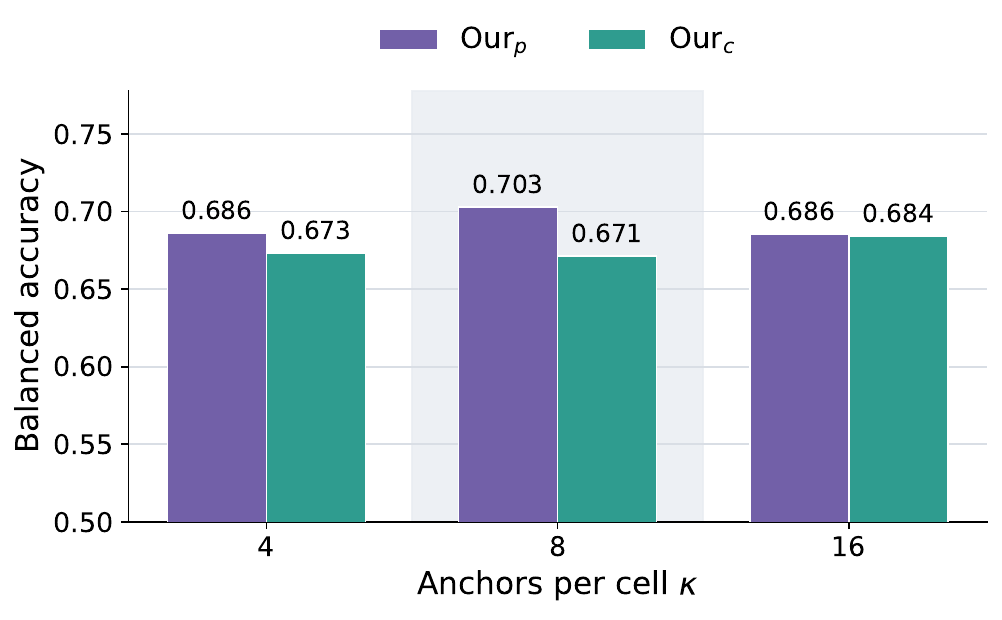}
\caption{Sensitivity to the number of anchors per cell $\kappa$ on FaithCoT-Bench using Qwen3. Results are macro-averaged over the four sources; shading marks the default setting $\kappa=8$.}
\label{fig:anchor-sensitivity}
\end{figure}

Figure~\ref{fig:anchor-sensitivity} shows modest sensitivity to $\kappa$. Our$_p$ performs best at $\kappa=8$, reaching 0.703 balanced accuracy, while Our$_c$ improves slightly at $\kappa=16$. We use $\kappa=8$ as the default because it provides the strongest Profile-based result without unnecessarily enlarging the anchor dictionary.

\end{document}